\documentclass{egpubl}

\usepackage{pg2022}

\SpecialIssuePaper         

\usepackage[T1]{fontenc}
\usepackage{dfadobe}  

\usepackage{cite}  
\BibtexOrBiblatex
\electronicVersion
\PrintedOrElectronic
\ifpdf \usepackage[pdftex]{graphicx} \pdfcompresslevel=9
\else \usepackage[dvips]{graphicx} \fi

\usepackage{egweblnk}

\usepackage{amsmath}
\usepackage{amsthm}
\usepackage{booktabs}
\usepackage{algorithm}
\usepackage{algorithmic}
\usepackage{amssymb}
\usepackage{multirow}

\title[ MODNet: Multi-offset Point Cloud Denoising Network Customized for Multi-scale Patches]%
      { MODNet: Multi-offset Point Cloud Denoising Network Customized for Multi-scale Patches}

\author[A. Huang, Q. Xie, Z. Wang, D. Lu, M. Wei \& J. Wang]
{\parbox{\textwidth}{
\centering A. Huang$^{1}$\orcid{0000-0001-7145-8535},
\centering Q. Xie$^{2}$,
\centering Z. Wang$^{1}$,
\centering D. Lu$^{3}$,
\centering M. Wei$^{1}$
        and J. Wang$^{1}$ \thanks{J. Wang is the corresponding author.}
        }
        \\
{\parbox{\textwidth}{\centering $^1$Nanjing University of Aeronautics and Astronautics, Nanjing, China\\
         $^2$University of Oxford, Oxford, UK\\
         $^3$University of Waterloo, Waterloo, Canada
       }
}
}

\begin{document}



\maketitle
\begin{abstract}

The intricacy of 3D surfaces often results cutting-edge point cloud denoising (PCD) models in surface degradation including remnant noise, wrongly-removed geometric details. Although using multi-scale patches to encode the geometry of a point has become the common wisdom in PCD, we find that simple aggregation of extracted multi-scale features can not adaptively utilize the appropriate scale information according to the geometric information around noisy points. 
It leads to surface degradation, especially for points close to edges and points on complex curved surfaces.
We raise an intriguing question -- if employing multi-scale geometric perception information to guide the network to utilize multi-scale information, can eliminate the severe surface degradation problem? 
To answer it, we propose a Multi-offset Denoising Network (MODNet) customized for multi-scale patches. 
First, we extract the low-level feature of three scales patches by patch feature encoders.
Second, a multi-scale perception module is designed to embed multi-scale geometric information for each scale feature and regress multi-scale weights to guide a multi-offset denoising displacement.
Third, a multi-offset decoder regresses three scale offsets, which are guided by the multi-scale weights to predict the final displacement by weighting them adaptively. 
Experiments demonstrate that our method achieves new state-of-the-art performance on both synthetic and real-scanned datasets. 
\textit{Our code is publicly available at \href{https://github.com/hay-001/MODNet}{https://github.com/hay-001/MODNet}}.


\end{abstract}  

\section{Introduction}

With the popularity of consumer-level 3D scanners, point clouds are widely used in research areas such as robotics~\cite{he2020pvn3d}, and 3D measurement~\cite{xie2020rrcnet}. Regrettably, raw point clouds scanned by 3D scanners are often noisy. 
Noisy point clouds highly affect its accuracy and characteristics, making point cloud denoising a vital step before further processing.

\begin{figure}
  \centering
  \includegraphics[width=\linewidth]{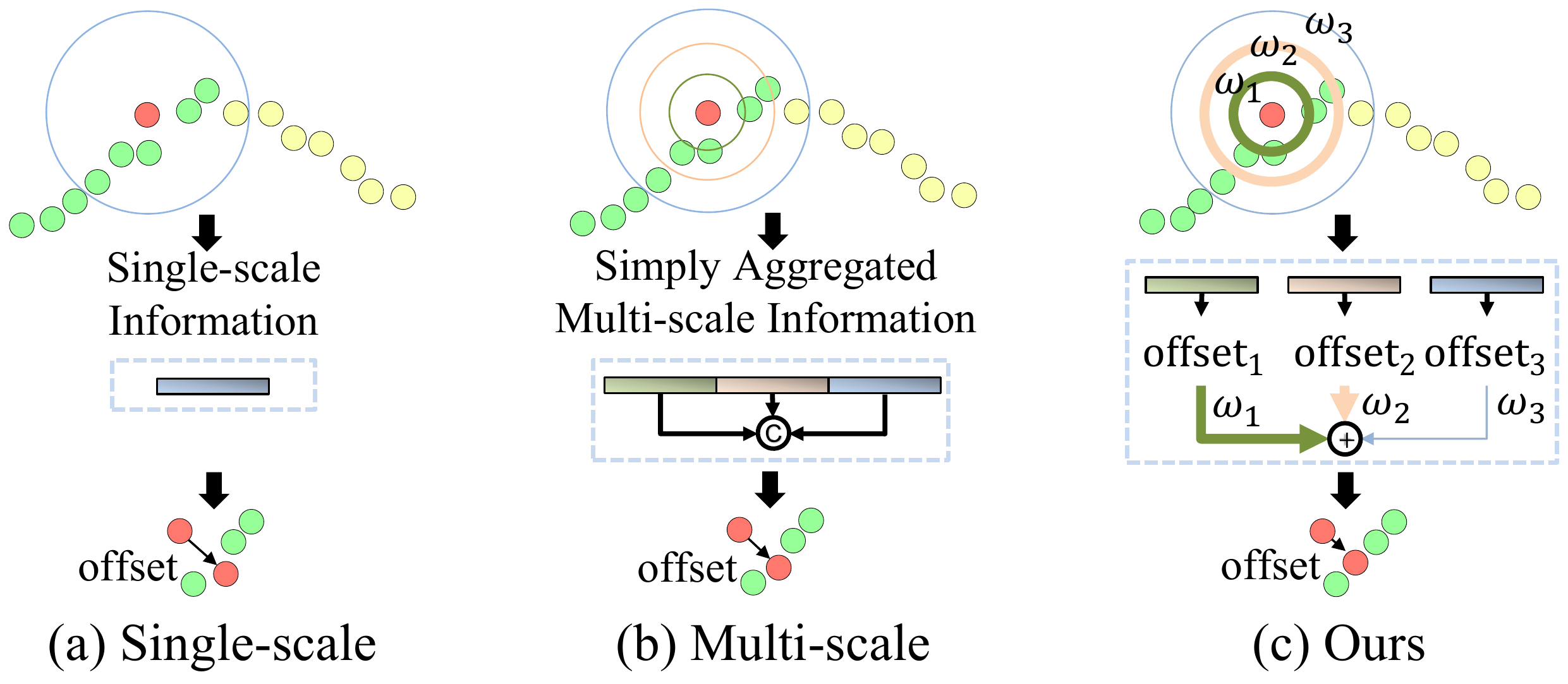}
  \caption{\label{fig:ms}Illustration of different ways to use patch information on denoising network. Different ring line widths indicate the weights in the utilization of various scale information on network according to different neighboring geometric information of central points (Red points). Yellow points are the negative effect points for denoising. $offset_{1}, offset_{2}$ and $offset_{3}$ are three different scale denoising offsets, which are guided by the multi-scale weights to predict the final $offset$ (i.e., displacement) by weighting them adaptively.}
\end{figure}

Existing point cloud denoising methods include optimization-based methods~\cite{alexa2001point,mattei2017point,zeng20193d,alexa2003computing,lipman2007parameterization,huang2009consolidation,sun2015denoising,hu2021dynamic} and deep learning-based methods~\cite{rakotosaona2020pointcleannet,roveri2018pointpronets,pistilli2020learning}. 
In recent years, the research of deep learning-based point cloud denoising expands substantially thanks to advanced neural network architectures proposed for point clouds~\cite{2017PointNet,qi2017pointnet++,thomas2019kpconv}. Among them, pioneering works such as PointCleanNet~\cite{rakotosaona2020pointcleannet}, TotalDn~\cite{hermosilla2019total} and Pointfilter~\cite{zhang2020pointfilter} are proposed.
Pointfilter and PointCleanNet take as input single-scale neighboring points of noisy points, as illustrated in Fig.~\ref{fig:ms} (a).
When the scale of the patch is not appropriate, this will introduce redundant neighborhood geometric information (leading to performance degradation) or lack sufficient neighborhood geometric information (leading to remnant noise).
It makes the network to be less robust to point clouds with various densities, noisy levels and complex curved surface structure. 
ECNet~\cite{yu2018ec} and TotalDn~\cite{hermosilla2019total} employ as input multi-scale neighboring points of noisy points, while only aggregating the extracted multi-scale features simply, as shown in Fig.~\ref{fig:ms} (b). 
It makes the network to be less robust to deal with redundant neighborhood geometric information (leading to performance degradation). 
Thus, none of these methods adaptively use the appropriate scale information according to the geometric information around noisy points, leading to underutilization of the multi-scale information.

To address this issue, we design a novel multi-offset denoising network, MODNet. Our network is an encoder-decoder-based architecture which straightforwardly takes as input three scale raw neighboring points of each noisy point. It predicts three different scale denoising offsets, which are guided by the multi-scale weights to predict the final denoising displacement, as shown in Fig.~\ref{fig:ms} (c).
MODNet is also the first denoising neural network customized for multi-scale patches input, which makes full use of the advantage of multi-scale information. 
Our MODNet consists of three parts: Patch Feature Encoder (PFE), Multi-Scale Perception Module (MSPM) and Multi-Offset Decoder (MOD). First, each patch feature encoder employs a PointNet-based feature extractor to extract the low-level features of each scale patch. 
Then, the outputs of patch feature encoders are fed into the multi-scale perception module for embedding multi-scale information for each scale low-level feature. 
And it regresses the multi-scale weights to guide multi-offset denoising. 
Finally, the multi-offset decoder predicts three scale denoising offsets, which are guided by the multi-scale weights to predict the final denoising displacement by weighting them adaptively.
Given a noisy point cloud as input, our MODNet can automatically and robustly predict a corresponding clean point cloud, by removing noise and preserving more geometric details. 
Extensive experiments demonstrate the denoising capability of our MODNet under a variety of noise models, better performance than the state-of-the-art techniques, in terms of both visual quality and error metrics. The main contributions of our work include:

\begin{itemize}
\item We propose a novel point cloud denoising network, which takes as input multi-scale patches of noisy points. 
MODNet, to the best of our knowledge, is the first denoising neural network customized for multi-scale patches input, which is able to adaptively utilize appropriate scale information of each point according to the multi-scale geometric perception information.
\item We design a multi-scale feature fusion perception module, named Multi-scale Perception Module (MSPM), for the efficient embedding of multi-scale information for each scale feature and guiding the multi-offset denoising displacement.
\item To make better use of the advantage of multi-scale information, we introduce a Multi-offset Decoder (MOD), which is capable of regressing three different scale denoising offsets. Finally, these denoising offsets are guided by the multi-scale weights to predict the final denoising displacement by weighting them adaptively.
\end{itemize}

\section{Related Work}
\label{related_work}

\subsection{Optimization-based denoising}

Early methods of 3D point cloud denoising are mainly based on optimization. These optimization-based methods design different geometric constraints for denoising and solve them through optimization methods. Existing optimization-based denoising methods can be divided into the following categories: local surface fitting methods, sparsity-based methods and graph-based methods.

The first category methods project noisy points onto the fitted surface. Alexa et al. ~\cite{alexa2001point} proposes a moving least squares (MLS) approach to calculate the optimal fitting surface of the point cloud. Its robust extensions~\cite{alexa2003computing, 2007Algebraic, oztireli2009feature} based on MLS are also widely used for point cloud denoising. Similarly, other surface fitting methods~\cite{cazals2005estimating,  lipman2007parameterization, huang2009consolidation, huang2013edge} have been proposed for point cloud denoising. Cazals et al. ~\cite{cazals2005estimating} proposed jet fitting, which encodes all local geometric quantities, such as normal direction and curvatures. Lipman et al.~\cite{lipman2007parameterization} proposed the locally optimal projection (LOP) to produce a set of points to describe the underlying surface. Huang et al.~\cite{huang2009consolidation, huang2013edge} further advanced this technique, e.g., weighted LOP (WLOP). whereas these methods tend to oversmooth the results, while removing the high-level noise.

Sparsity-based denoising methods are based on the sparse representation theory ~\cite{avron2010, sun2015denoising, mattei2017point}. These methods mainly optimize the coordinates of points by reconstructing normals. Mattei et al.~\cite{mattei2017point} proposed Moving Robust Principal Components Analysis (MRPCA) to restore the noisy point cloud without oriented normals. Whereas, These methods are less robust to high-level noisy point cloud.

Finally, graph-based denoising methods are based on the graph signal processing theory~\cite{shuman2013emerging}. These techniques~\cite{schoenenberger2015graph,hu2021graph,hu2020feature,hu2021dynamic} mainly employ graph filters to denoise the point cloud represented by graphs. Zeng et al.~\cite{zeng20193d} proposed patch-based graph Laplacian regularization (GLR) for denoising. However, the denoising performance of graph-based denoising methods for high-level noisy point cloud is still unstable. 

All these optimization-based methods are mainly based on geometric priors, so there are often difficulties in balancing denoising performance and preserving sharp details.

\subsection{Deep learning-based denoising}

In recent years, the research of deep learning-based point cloud denoising expands substantially thanks to advanced neural network architectures proposed for point clouds~\cite{2017PointNet,qi2017pointnet++,lu2022transformers}.
Existing deep learning-based denoising methods mainly take as input a single-scale patch to encode the local geometric information and predict the displacement and apply the inverse displacement to each noisy point.
For instance, PointCleanNet(PCN)~\cite{rakotosaona2020pointcleannet}, similar to the framework of PCPNet~\cite{guerrero2018pcpnet}, was proposed to estimate denoising displacement with a single patch as input and remove outliers separately in the noisy point cloud. 
Zhang et al.~\cite{zhang2020pointfilter} proposed the Pointfilter, a point-wise learning encoder-decoder-based framework to take as input a patch. And it learns a latent displacement supervised by a loss function sensitive to sharp features.
However, above methods are less robust to denoise point clouds with various densities and noisy levels.
Recently, multi-scale neighboring information is demonstrated to be critical for point cloud denoising by several multi-scale patches based methods. 
Yu et al.~\cite{yu2018ec} presented an edge-aware point cloud consolidation network to achieve denoising trivial low-level noise, which takes PointNet++~\cite{qi2017pointnet++} for multi-scale feature embeding. But it cannot well preserve tiny features of surfaces. 
For achieving unsupervised deep learning-based denoising, TotalDn~\cite{hermosilla2019total} was proposed.
In TotalDn, it introduces a spatial prior term, that steers converges to the unique closest out of the many possible modes on a manifold by using two scale information. However, denoising results of TotalDn shrink dramatically.
Regretfully, these multi-scale patches based methods just simply aggregate the extracted multi-scale features.
To summarize, none of these methods adaptively use the appropriate scale information according to the geometric information around noisy points, leading to underutilization of the multi-scale information.

Furthermore, some other neural network architectures were proposed for denoising, which use CNN and graph convolution to achieve point cloud denoising.
Roveri et al.~\cite{roveri2018pointpronets} proposed the PointProNets, a CNN-based framework to consolidate high-quality point cloud. 
Francesca et al.~\cite{pistilli2020learning} proposed GPDNet, a graph convolutional network, to achieve robust denoising.

\section{Approach}
\label{method}

\begin{figure}[!t]
  \centering
  \includegraphics[width=\linewidth]{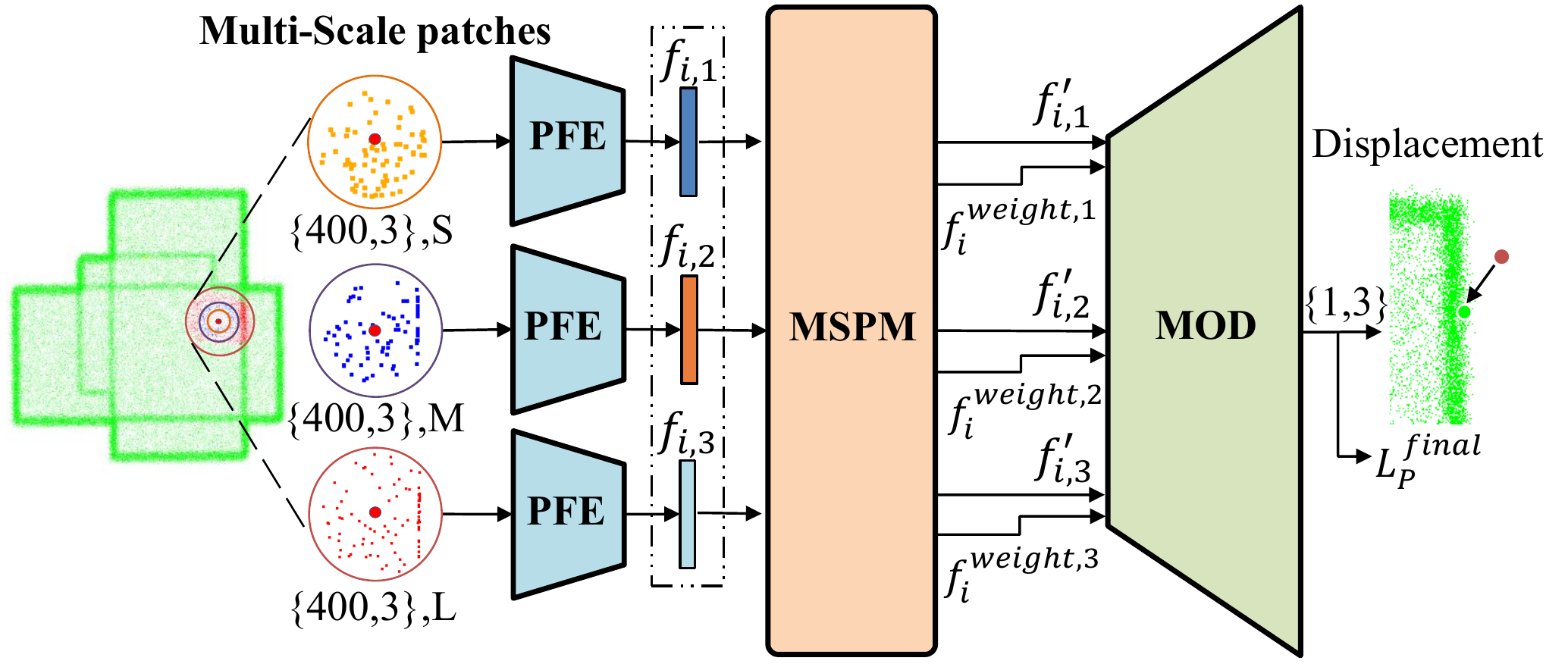}
  \caption{\label{fig:modnet}The architecture of MODNet. The input of MODNet are multi-scale patches of a noisy point. Then, the PFE (Section 3.3) extracts the low-level features of each scale patch (indicated by blue, orange and light blue). The MSPM (Section 3.4) is used to embed multi-scale information into each scale low-level feature and regress multi-scale weights for MOD. The MOD (Section 3.5) predicts the final displacement.}
\end{figure}

\begin{figure}[!t]
  \centering
  \includegraphics[width=\linewidth]{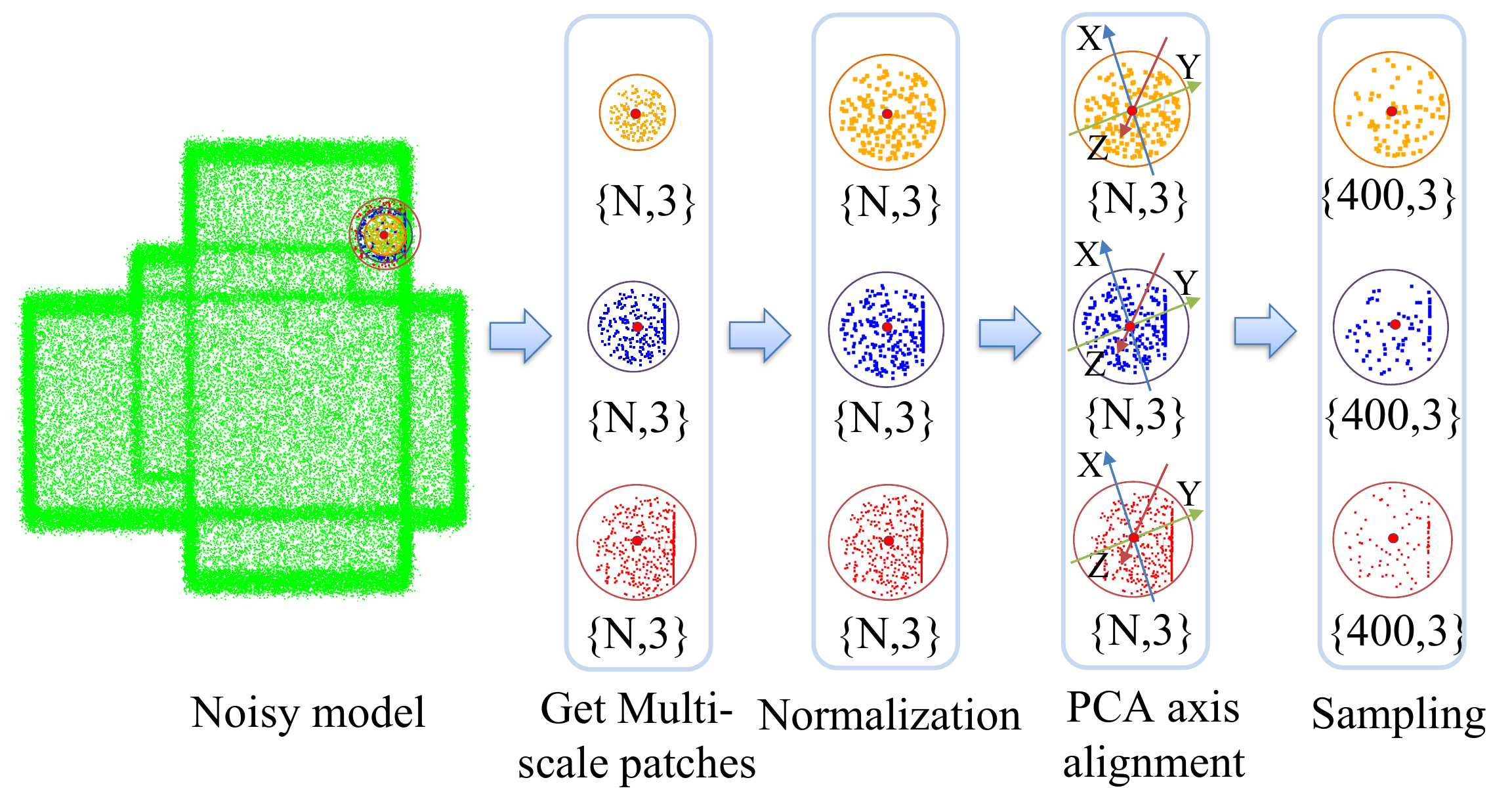}
  \caption{\label{fig:preprocessing}The preprocessing of multi-scale patches.}
\end{figure}

\subsection{Overview}
Given a noisy point cloud, we aim to recover the underlying noise-free point cloud by removing the additive noise. We formulate a noisy point cloud intuitively as follows:
\begin{equation}
\hat{P}=P+N, 
\end{equation}
where $\hat{P}=\left \{  \hat{p_{1}}, \hat{p_{2}}, .., \hat{p_{n}}\right |\hat{p_{n}}\in \mathbb{R}^{3}\}$ is a noisy point cloud, ${P}$ is the corresponding clean point cloud. $N$ is the additive noise. 

We solve this denoising problem in a local way, which means denoising a noisy point only depends on its neighboring points (patch). Specifically, we employ as input multi-scale neighboring points of a noisy point, defined as follows:

\begin{equation}
\begin{aligned}
 \hat{\wp_{i}}=\left \{ {\hat{\wp_{i, 1}}}, ..,{\hat{\wp_{i, k}}},..,{\hat{\wp_{i, K}}} \right \}, \\
\wp_{i}=\left \{ {\wp_{i, 1}}, ..,{\wp_{i, k}},..,{\wp_{i, K}} \right \},
\end{aligned}
\end{equation}

\begin{equation}
\begin{aligned}
 \hat{\wp_{i, k}}=\left \{ \hat{p_{j, k}}\mid \left \| \hat{p_{j, k}}-\hat{p_{i }} \right \|  \right < r_{k} \},\\
\wp_{i, k}=\left \{{p_{j, k}}\mid \left \| {p_{j, k}}-\hat{p_{i }} \right \|  \right < r_{k} \},
\end{aligned}
\end{equation}
where $\hat{p_{i}}, \hat{p_{j, k}}\in \hat{P}$, $p_{j, k} \in {P}$ and ${\hat{\wp_{i}}}$ is a set of noisy multi-scale patches ${\hat{\wp_{i, k}}}$ of point $\hat{p_{i}}$ and those corresponding ground truth multi-scale patches ${\wp_{i}}$ from a pair of point clouds ${\hat{P}}$ and ${P}$. $r_{k}$ is the patch radius of scale $k$.

Similar to PointCleanNet~\cite{rakotosaona2020pointcleannet}, we aim to learn the displacement for each point, rather than directly learning clean point cloud. Therefore, we formulate the denoising model:
\begin{equation}
\dot{p_{i}}= {\hat{p_{i}}}+f\left ( \hat{\wp_{i}}\right ),
\end{equation}
where $f\left (\cdot  \right )$ represents our MODNet, and $\dot{p_{i}}$ is the filtered point of the noisy  point ${\hat{p_{i}}}$.

The overview of our MODNet is depicted in Fig.~\ref{fig:modnet}, consisting of three parts: Patch Feature Encoder (PFE), Muliti-Scale Perception Module (MSPM) and Multi-Offset Decoder (MOD). First, each patch feature encoder employs a PointNet-based feature extractors to extract the low-level features of each scale patch. Then, the outputs of PFE are fed into the multi-scale perception module for embedding multi-scale information for each scale low-level feature.
And it regresses the multi-scale weights to guide multi-offset denoising. Finally, the multi-offset decoder predicts three different scale denoising offsets. Then, These three scale denoising offsets are guided by the multi-scale weights to predict the final denoising displacement by weighting them adaptively.

\subsection{Multi-scale Patches Preprocessing}

In order to improve the robustness of denoising, it is necessary to preprocess the multi-scale patches. The preprocessing of multi-scale patches similar to ~\cite{zhang2020pointfilter} is shown in Fig~\ref{fig:preprocessing}. The first step is to normalize the multi-scale patches, i.e., $\hat{\wp_{i}}=\left ( \hat{\wp_{i}}-\hat{p_{i}} \right) /r_{max}, \wp_{i}=\left ( \wp_{i}-\hat{p_{i}} \right) /r_{max}$, where $r_{max}$ is radius of the max scale. For rotation invariance, we align each patch by aligning its principle axes of the PCA with the Cartesian space. More specifically, the z-axis and x-axis of the patches are aligned with the last and the second principle axis respectively. Finally, to fix the input of the MOD-Net, we set the number of scales $K=3$, and the point number of patch $\left | \hat{p_{i,k}} \right | =400$. We pad the origin for patches with less than 400 points and randomly downsample the patches with more than 400 points. We set the radius $r_{k}$ of the three patches to $3\%$, $4\%$ and $5\%$ of the model’s bounding box diagonal length.

\subsection{Patch Feature Encoder}

Given three patches of different scales, the goal of three patch feature encoders is to extract the low-level features of each patch. For the backbone with patch as input, the encoder based on Pointnet~\cite{2017PointNet} shows impressive performance compared to other backbones~\cite{zhang2020pointfilter}.
We use a set of multi-layer perceptrons (MLP) with batch normalization and Relu function to encode each point of patch into per point feature. 
Then, we employ max pooling layers to obtain three low-level features $f_{i, 1}$, $f_{i, 2}$ and $f_{i, 3}$.

\subsection{Multi-scale Perception Module}

\begin{figure}[!t]
  \centering
  \includegraphics[width=\linewidth]{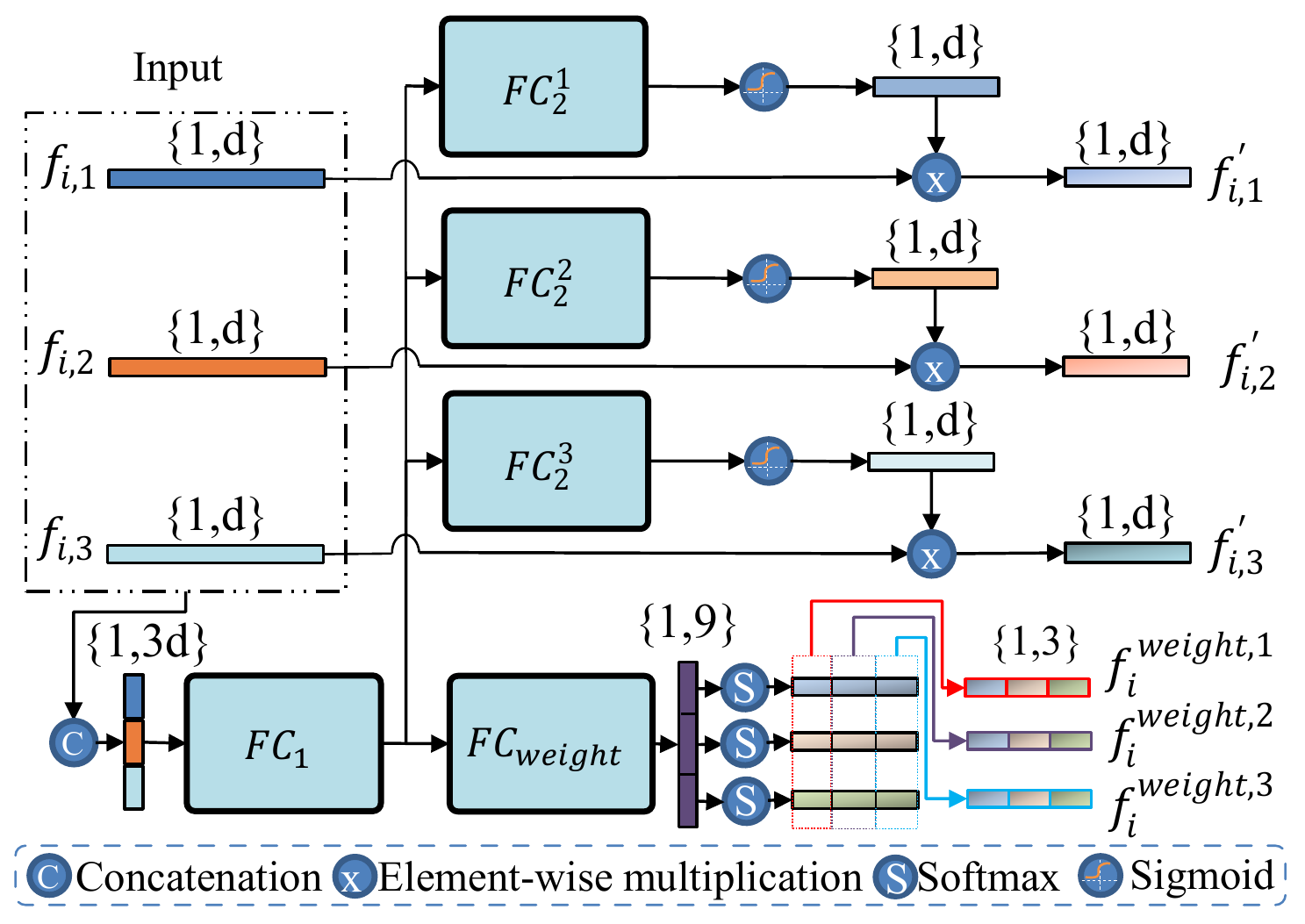}
  \caption{\label{fig:gspm}Illustration of MSPM in MODNet. Our method employs an attention mechanism to embed multi-scale information into each scale low-level feature and regress multi-scale weights for MOD. }
\end{figure}

A single-scale patch as input has some limitations in perceiving the geometric neighborhood information required for denoising. 
The excessively small scale has an insufficient perception of the geometric neighborhood information required for denoising, while the excessively large scale may introduce redundant geometric neighborhood information, resulting in performance degradation.
Therefore, our MODNet takes as input multi-scale patches. Multi-scale patches provide more geometric neighborhood information and the choice of adaptively selecting the optimal scale information.

First, the multi-scale perception module(MSPM) aggregates the outputs of the three encoders to obtain a multi-scale fusion feature as shown in Fig.~\ref{fig:gspm}. 
The multi-scale fusion feature can fully perceive the neighborhood geometric structure information around the $\hat{p_{i}}$.
Therefore, we call the fused vector as multi-scale feature vector. It can be expressed as follows:
\begin{equation}
 f_{i}=Concat \left ( f_{i, 1}, f_{i, 2}, f_{i, 3} \right ),  
\end{equation}
where $f_{i}\in \mathbb{R} ^{1\times 1536}$ is the multi-scale feature vector, and $Concat\left ( \cdot  \right )$ is the concatenation operation.

Then, the MSPM adopts the attention mechanism to embed the multi-scale feature to each scale low-level feature, which can be expressed as follows:
\begin{equation}
f_{i, k}^{'}= f_{i, k} \otimes Sig\left ( FC_{2}^{k}\left ( FC_{1}\left ( f_{i} \right ) \right ) \right ),
\end{equation}
where $ FC_{1}\left ( \cdot  \right )$, $ FC_{2}^{k}\left ( \cdot  \right )$ are fully connected layers, and $Sig\left ( \cdot  \right )$ denotes the $Sigmoid$ function. $\otimes$ is the element-wise multiplication. 

Meanwhile, the multi-scale perception module regresses three multi-scale weights ($1\times 3$) to guide the final multi-offset displacement, which can be expressed as follows:
\begin{equation}
\left[ f_{i}^{'x}, f_{i}^{'y}, f_{i}^{'z} \right]
= RP\left (FC_{weight}\left ( FC_{1}\left ( f_{i} \right ) \right )\right ),
\end{equation}
\begin{equation}
\begin{aligned}
&\left [ f_{i}^{weight,1},f_{i}^{weigh,2},f_{i}^{weigh,3} \right ]=\\
&\left [ Soft\left ( f_{i}^{'x}  \right ),Soft\left ( f_{i}^{'y}  \right ),Soft\left ( f_{i}^{'z}  \right ) \right ]^{T},
\end{aligned}
\end{equation}
where $FC_{weight}\left ( \cdot  \right )$ is two-layer fully connected layers (FC), $Soft\left ( \cdot  \right )$ denotes the $Softmax$ function, and $RP\left ( \cdot  \right )$ converts the vector ($1\times 9$) to a weight matrix ($3\times 3$).

\subsection{Multi-offset Decoder}

\begin{figure}[!t]
  \centering
  \includegraphics[width=\linewidth]{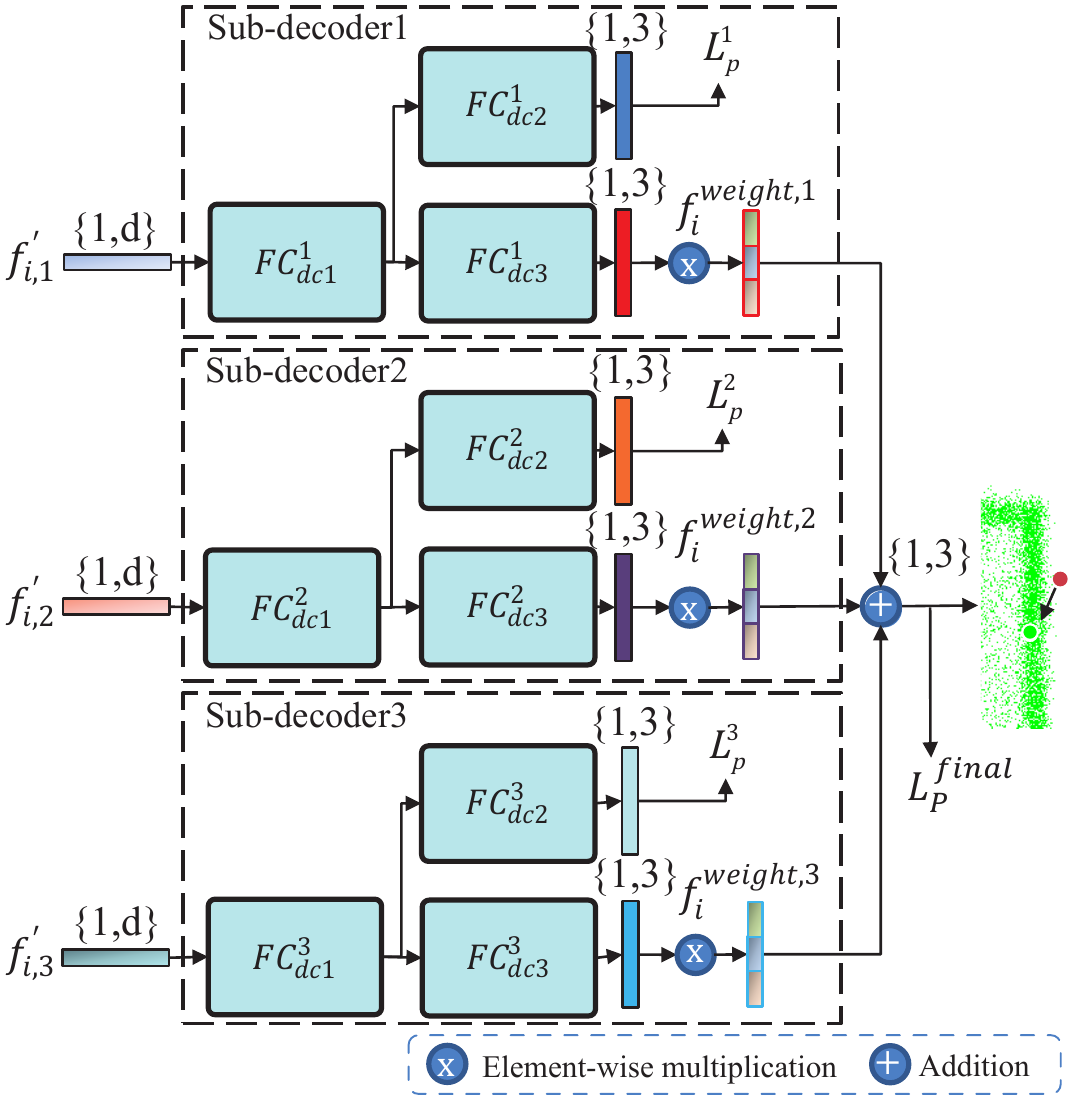}
  \caption{\label{fig:mod}The structure of Multi-offset Decoder (MOD) in MODNet. It regresses three different scale denoising offsets, which are guided by the multi-scale weights to predict the final denoising displacement by weighting them adaptively. }
\end{figure}

For noisy points with different geometric information around these points, the suitable scale information is not universal.
Therefore, our method tends to adaptively use the appropriate scale information according to the perceived multi-scale information around noisy points, as illustrated in Fig.~\ref{fig:ms}. 
Instead of directly predicting the denoising displacement like existing denoising network, we propose a multi-offset decoder, as shown in Fig.~\ref{fig:mod}. It can better exploit the advantages of multi-scale information. 
First, multi-offset decoder employs three sub-decoders composed of a series of fully connected layers to predict denoising offsets under three different scale patches. The sub-decoder can be formulated as follows:
\begin{equation}
f_{i, k}^{''}= FC_{dc1}^{k}\left (f_{i, k}^{'}\right ),
\end{equation}
\begin{equation}
dp_{i,k}^{pre}= FC_{dc2}^{k}\left (f_{i, k}^{''}\right ), 
\end{equation}
\begin{equation}
offset_{i,k}= FC_{dc3}^{k}\left (f_{i, k}^{''}\right ), 
\end{equation}
where $ Fs_{dc1}^{k}\left ( \cdot  \right ), FC_{dc2}^{k}\left ( \cdot  \right )$ and $FC_{dc2}^{k}\left ( \cdot  \right )$ are fully connected layers (FC).  $offset_{i,k}$ is $i$-th point offset of $k$-th scale. And three single scale denoising offsets $dp_{i}^{pre}$ are supervised by three different loss functions. Specifically, these denoising offsets are not used for final denoising, but to lead sub-decoders of different scales to focus on learning the key feature of denoising at their respective scales. The specific introduction is shown in Section 3.6.
Then these three different scale denoising offsets are guided by the multi-scale weights to predict the final denoising displacement by weighting them adaptively. It can be expressed as follows:

\begin{equation}
\begin{aligned}
dp_{i}=offset_{i,1} \otimes f_{i}^{weight,1}&+offset_{i,2} \otimes f_{i}^{weight,2}\\&+offset_{3} \otimes f_{i}^{weight,3},
\end{aligned}
\end{equation}
where $dp_{i}$ is the final denoising displacement of $i$-th noisy point. $\otimes$ is the element-wise multiplication.

\subsection{Loss}

For the performance of point cloud denoising neural network, the choice of loss is critical. The L2 distance which has been used in PointCleanNet~\cite{rakotosaona2020pointcleannet} is the general loss for point clouds denoising. Inspired by pointfilter~\cite{zhang2020pointfilter}, the loss function $L_{p}$ for single scale patch consists of a projection loss function $L_{s}^{k}$ and a repulsion term $L_{r}^{k}$. The function $L_{s}^{k}$ preserves sharp features by considering the distance and the normal similarity between the current denoising point $\hat{p_{i}}$ and its neighboring points of ground truth patch. The projection loss function $L_{s}^{k}$ takes the form:
\begin{equation}
 L_{s}^{k}=\frac{\sum_{p_{j, k}\in \wp_{i, k}}^{} \left| \left (\dot{p_{i}}-p_{j, k} \right )\cdot n_{p_{j, k}}^{T}\right | \cdot \phi_{i}^{k}\cdot \theta_{i}^{k}}{\sum_{p_{j, k}\in \wp_{i, k}}^{}\phi_{i}^{k}\cdot \theta_{i}^{k}}, 
\end{equation}

\begin{equation}
\phi_{i}^{k}= \exp \left ( - \frac{\left  \| \dot{p_{i}}-p_{j,k}   \right \|^2}{\varepsilon_{p}^2 }  \right ), 
\end{equation}

\begin{equation}
\theta_{i}^{k}= \exp \left ( - \frac{1-n_{\dot{p_{i}}}^{T}\cdot n_{p_{j,k}}}{1-\cos\left ( \varepsilon _{n} \right )   }  \right ), 
\end{equation}
where $\dot{p_{i}}$ is the filtered point of the noisy point $\hat{p_{i}}$, and $n_{p_{j, k}}$ is the ground-truth normal of the point $p_{j, k}$. $\phi \left (\left  \| \dot{p_{i}}  -p_{j, k} \right\| \right)$ is a Gaussian function giving larger weights to the points near $\dot{p_{i}}$. And $ \theta  \left ( n_{\dot{p_{i}}}, n_{p_{j}}  \right )$ is a function giving larger weights to the neighboring points in the patch with more similar normal to $\dot{p_{i}}$. $\varepsilon_{p}$ is defined as $\varepsilon_{p}=4\sqrt{dobb/m}$ and $dobb$ is the length of the diagonal of the bounding box of patch $\wp_{i, k}$.
The $m$ denotes the point number of ground truth patch. 
$\varepsilon_{n}$ is the support angle ($15^{\circ}$ by default). Furthermore, the repulsion term $L_{r}^{k}$ makes the denoising points be distributed uniformly. It is defined as
\begin{equation}
L_{r}^{k}= \max_{p_{j}\in \wp_{i}} \left | \dot{p_{i}}-p_{j} \right |, 
\end{equation}
Overall, we formulate the loss function $L_{p}$ as:
\begin{equation}
L_{p}=\alpha L_{s}^{k}+\left ( 1-\alpha \right ) L_{r}^{k}, 
\end{equation}
where $\alpha$ is a trade-off parameter to control the repulsion force, and we empirically set  $\alpha=0.97$ in our training stage.

Then the loss function $L_{dp}$ of three scales is formulated as:
\begin{equation}
L_{dp}=L_{p}^{1}+L_{p}^{2}+L_{p}^{3},
\end{equation}
where $L_{p}^{1}, L_{p}^{2}$ and $L_{p}^{3}$ respectively denote the loss function $L_{p}$ for offsets of three different scale. Finally, we combine all of losses:
\begin{equation}
L=\beta \cdot L_{dp}+  L_{p}^{final},
\end{equation}
where $\beta$ is a trade-off parameter to balance the different losses, and we set  $\beta=0.2$ in our training stage. The $L_{p}^{final}$ denotes the projection loss function of the final denoising displacement and we set the value of ground truth $\left | \wp_{i,final} \right | =500, r_{final}=5\%$. These hyperparameters are obtained according to the experimental verification.

\section{Results and Discussions}
\label{result}

\begin{table*}[]
\centering
\begin{tabular}{c|llrrrrrrrrr}
\hline
\multirow{2}{*}{Metrics} & \multicolumn{1}{c}{\multirow{2}{*}{Model}} & \multicolumn{3}{c}{0.5\%} & \multicolumn{3}{c}{1\%} & \multicolumn{3}{c}{1.5\%} & \multicolumn{1}{c}{\multirow{2}{*}{Average}} \\
 &  & \multicolumn{1}{c}{10K} & \multicolumn{1}{c}{20K} & \multicolumn{1}{c}{50K} & \multicolumn{1}{c}{10K} & \multicolumn{1}{c}{20K} & \multicolumn{1}{c}{50K} & \multicolumn{1}{c}{10K} & \multicolumn{1}{c}{20K} & \multicolumn{1}{c}{50K} & \multicolumn{1}{c}{} \\ \hline
\multirow{6}{*}{CD$(10^{-4})$} 
 & WLOP~\cite{huang2009consolidation} & 8.68 & 6.33 & 4.19 & 13.2 & 7.26 & 7.04 & 19.0 & 10.7 & 12.55 & 9.88 \\
 & ECN~\cite{yu2018ec} & 5.12 & 2.85 & 1.29 & 6.40 & 3.74 & 2.08 & 7.90 & 5.14 & 4.11 & 4.29 \\
 & PCN~\cite{rakotosaona2020pointcleannet} & 3.39 & 2.39 & 1.22 & 7.33 & 4.09 & 1.62 & 11.5 & 6.32 & 2.09 & 4.44 \\
 & GPD~\cite{pistilli2020learning} & 3.39 & 2.43 & 1.44 & 4.87 & 2.84 & 1.45 & 6.23 & 3.64 & 1.99 & 3.14 \\
 & PF~\cite{zhang2020pointfilter} & 2.90 &  1.84 & \textbf{0.92} & 5.17 & 2.86 & 1.31 & 7.40 & 4.11 & 1.93 & 3.16 \\ \cline{2-12} 
 & Ours & \textbf{2.73} & \textbf{1.83} & 0.93 & \textbf{4.68} & \textbf{2.64} & \textbf{1.26} & \textbf{5.96} & \textbf{3.31} & \textbf{1.63} & \textbf{2.78} \\ \hline\hline
\multirow{6}{*}{MSE$(10^{-2})$} 
& WLOP~\cite{huang2009consolidation} & 4.22 & 2.72 & 2.23 & 4.45 & 2.89 & 2.67 & 4.88 & 3.32 & 3.34 & 3.41 \\
 & ECN~\cite{yu2018ec} & 3.58 & 2.54 & 1.64 & 3.64 & 2.65 & 1.79 & 3.76 & 2.82 & 2.16 & 2.73 \\
 & PCN~\cite{rakotosaona2020pointcleannet} & 3.51 & 2.49 & 1.59 & 3.70 & 2.63 & 1.65 & 4.04 & 2.85 & 1.72 & 2.69 \\
 & GPD~\cite{pistilli2020learning} & 3.53 & 2.55 & 1.69 & 3.47 & 2.53 & 1.69 & 3.41 & 2.53 & 1.69 & 2.57 \\
 & PF~\cite{zhang2020pointfilter} & 3.41 & 2.41 & 1.55 & 3.44 & 2.46 & 1.59 & 3.59 & 2.58 & 1.69 & 2.52 \\ \cline{2-12} 
 & Ours & \textbf{3.40} & \textbf{2.40} & \textbf{1.54} & \textbf{3.39} & \textbf{2.44} & \textbf{1.58} & \textbf{3.46} & \textbf{2.50} & \textbf{1.64} & \textbf{2.48} \\ \hline\hline
\multirow{6}{*}{P2M$(10^{-4})$} 
 & WLOP~\cite{huang2009consolidation} & 2.88 & 2.66 & 1.394 & 4.79 & 3.19 & 2.49 & 7.71 & 5.25 & 4.92 & 3.92 \\
 & ECN~\cite{yu2018ec} & 1.30 & 0.68 & 0.29 & 1.84 & 1.07 & 0.63 & 2.54 & 1.70 & 1.49 & 1.28 \\
 & PCN~\cite{rakotosaona2020pointcleannet}  & 1.04 & 0.59 & 0.28 & 2.35 & 1.24 & 0.44 & 4.30 & 2.29 & 0.65 & 1.46 \\
 & GPD~\cite{pistilli2020learning} & 1.07 & 0.62 & 0.25 & 2.18 & 1.47 & 0.58 & 2.52 & 1.31 & 0.78 & 1.20 \\
 & PF~\cite{zhang2020pointfilter} & 0.77 & \textbf{0.37} & \textbf{0.16} & 1.41 & 0.72 & 0.32 & 2.39 & 1.28 & 0.58 & 0.89 \\ \cline{2-12} 
 & Ours & \textbf{0.71} & 0.38 & 0.17 & \textbf{1.20} & \textbf{0.64} & \textbf{0.31} & \textbf{1.78} & \textbf{0.97} & \textbf{0.49} & \textbf{0.74} \\ \hline
\end{tabular}
\caption{\label{tab:Quantitative}Quantitative comparisons by testing various methods on synthetic noisy models with different noise levels and point densities.}
\end{table*}

\subsection{Datasets}
The training dataset consists of 11 CAD clean models and 11 clean non-CAD models provided by the pointfilter~\cite{zhang2020pointfilter}. 
And we sample 10000, 20000 and 50000 points on each clean model. 
Considering that the noise of depth cameras, like Kinect, is similar to Gaussian noise~\cite{nguyen2012modeling}, we assume noise as Gaussian distribution like most methods. Based on this, each clean model corresponds to 6 synthesized noisy models with 6 standard deviations of the clean model’s bounding box diagonal length ($0.0\%$ to $1.5\%$) Gaussian noise. Besides, the normal information for clean models is only required in training phase.

To verify the performance of our model, we perform it on synthetic models which contain 10000, 20000 and 50000 points randomly sampled from its original surface. Similar to the training set, the synthetic models consist of 96 clean models and 288 synthetic noisy models with 3 standard deviations of the clean model’s bounding box diagonal length ($0.5\%$, $ 1\%$ and $1.5\%$) Gaussian noise. 
Moreover, we test a real-world dataset Paris-rue-Madame~\cite{serna2014paris} to verify the effectiveness of the method in some real scenarios.

\subsection{Training details}
Our MODNet is implemented using PyTorch. Our model is trained on a PC with a single Nvidia GeForce RTX 2070s GPU (8GB memory). We employ the SGD optimizer to train our MODNet with a batch size of 200. We set 40 training epochs and 1e-4 initial learning rate. And the learning rate decreases from 1e-4 to 1e-7 with increasing epochs. The training of MODNet takes about 12 hours.

\subsection{Evaluation Metric}
To evaluate the denoising performance of different methods, we adopted the Chamfer Distance (CD) to evaluate the "global similarity" between the filtered point cloud and its corresponding ground truth. The Chamfer Distance metric mutually detects the nearest neighbor in the other point cloud and sums the squared distances up. The CD can be expressed as:
\begin{equation}
\begin{aligned}
CD(\dot{P}, P)=&\frac{1}{M}\sum_{p_{i}\in P }^{} \min_{\dot{p_{j}}\in \dot{P}}\left ( \left \| p_{i}-\dot{p_{j}} \right \|   _{2}^{2}\right)\\
&+\frac{1}{\dot{M}}\sum_{\dot{p_{j}}\in \dot{P} }^{} \min_{p_{i}\in P}\left ( \left \| \dot{p_{j}}-p_{i} \right \| _{2}^{2}\right  ) ,  \\
\end{aligned}
\end{equation}
Here $\dot{P}, P$ are the filtered point cloud and its corresponding ground truth point cloud. $\dot{M}$ and $M$ are the cardinalities of the filtered $\dot{P}$ and ground truth $P$ point clouds. 

Moreover, to evaluate the "local similarity" between the filtered point cloud and its corresponding ground truth, we employ the mean square distance error (MSE)~\cite{lu2017gpf}. The MSE metric evaluates the distance between each point in ground truth and its closest points in the filtered point cloud. It takes the form: 
\begin{equation}
D(p_{i})=\frac{1}{N}\sum_{ \dot{p_{j}}\in NN\left ( p_{i} \right ) }\left\| p_{i}-\dot{p_{j}} \right \|  _{2}^{2}  
\end{equation}
where $p_{i}$ is the ground truth point and $\dot{p_{j}}$ is one of its neighboring point in the filtered point cloud. $NN(*)$ represents the nearest neighbors, and $N=\left | NN(p_{i}) \right | $. we set $N$ to 10. 

Besides the two above metrics, we also introduce the point-to-mesh (P2M)~\cite{ravi2020accelerating} to evaluate the error between the filtered point cloud and its corresponding ground truth mesh.

\begin{table}[]
\centering
\begin{tabular}{@{}lrrr@{}}
\toprule
\multirow{2}{*}{Model} & \multicolumn{3}{c}{Runtime(s)} \\
 & 10K & 20K & 50K \\ \midrule
WLOP & 2.21 & 4.51 & 11.2 \\
ECN & 1.93 & 3.87 & 9.28 \\
GPD & 21.3 & 43.2 & 99.5 \\
PCN & 55.0 & 103 & 245 \\
PF & 5.50 & 10.2 & 24.6 \\ \hline
Ours & 16.3 & 32.4 & 81.1 \\ \bottomrule
\end{tabular}
 \caption{\label{tab:Runtime}Runtime performance (in seconds) for different approaches on three test subsets with different number of points. All examples were run on the same computer configurations.}
\end{table}

 \begin{figure*}[!t]
  \centering
  \includegraphics[width=1\textwidth]{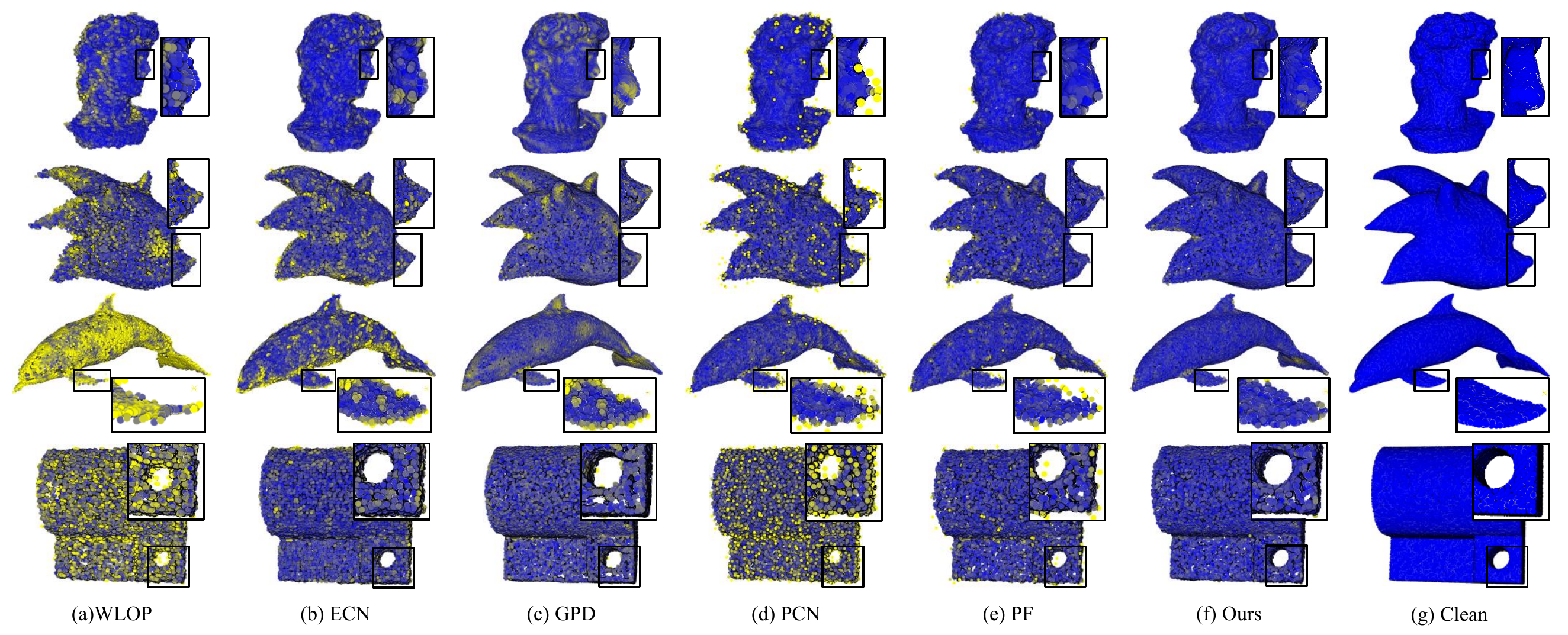}
  \caption{ \label{fig:Visual other methods}Visual comparisons of denoising methods under Gaussian noise. Yellower Points are farther away from the ground truth surface. Obviously, our method outperforms others in terms of both noise removal and feature preservation; see particularly the blown-up views. }
\end{figure*}

\subsection{Quantitative Comparison}

\textbf{Errors}. We compare our MODNet quantitatively to the state-of-the-art denoising methods, including WLOP~\cite{huang2009consolidation}, ECN~\cite{yu2018ec}, GPD~\cite{pistilli2020learning}, PCN~\cite{rakotosaona2020pointcleannet} and PF~\cite{zhang2020pointfilter}. We first calculate the errors over the test set, in terms of the Chamfer distance (CD), mean square error (MSE) and point-to-mesh distance (P2M). As presented in Table~\ref{tab:Quantitative}, our MODNet significantly outperforms previous existing methods. In particular, our method still has the promising performance whether input is low-density point cloud with 10000 points lacking local geometric information or point cloud with a $1.5\%$ high-level noise.

\textbf{Runtime}.Table~\ref{tab:Runtime} summaries the runtime of each method on test set. As for optimization-based methods, the WLOP is fast than most of deep learning-based methods. However, in order to acquire better results, WLOP needs multi-step preprocessing and trial-and-error efforts to tune parameters, which take a much longer extra time in practice. On the contrary, ECN is the fastest method among the deep-learning-based methods(GPD, PCN, PF and ours) and even faster than optimized-based methods(WLOP). And PCN is the slowest method due to its complex backbone structure. Our MODNet is faster than PCN and GPD, but slower than PF. This is because our MODNet requires to search multi-scale patches for each noisy point and extract the feature of each scale's patch. 

\textbf{Robustness}.
To test the generalizability of our MODNet, We use a variety of other noise models to test our models, including discrete noise (DN), Laplace noise (LN) and uniform noise (UN). Moreover, the performance of our model is then evaluated using 3$\%$ high-level Gaussian noise (HLN). From Table.~\ref{tab:Quantitative_rebust}, we can observe that our MODNet achieves superior performance of generalizability for other noise models and high-level noise. As such, our MODNet is more robust than other compared methods.

 \begin{figure*}[!t]
  \centering
  \includegraphics[width=1\textwidth]{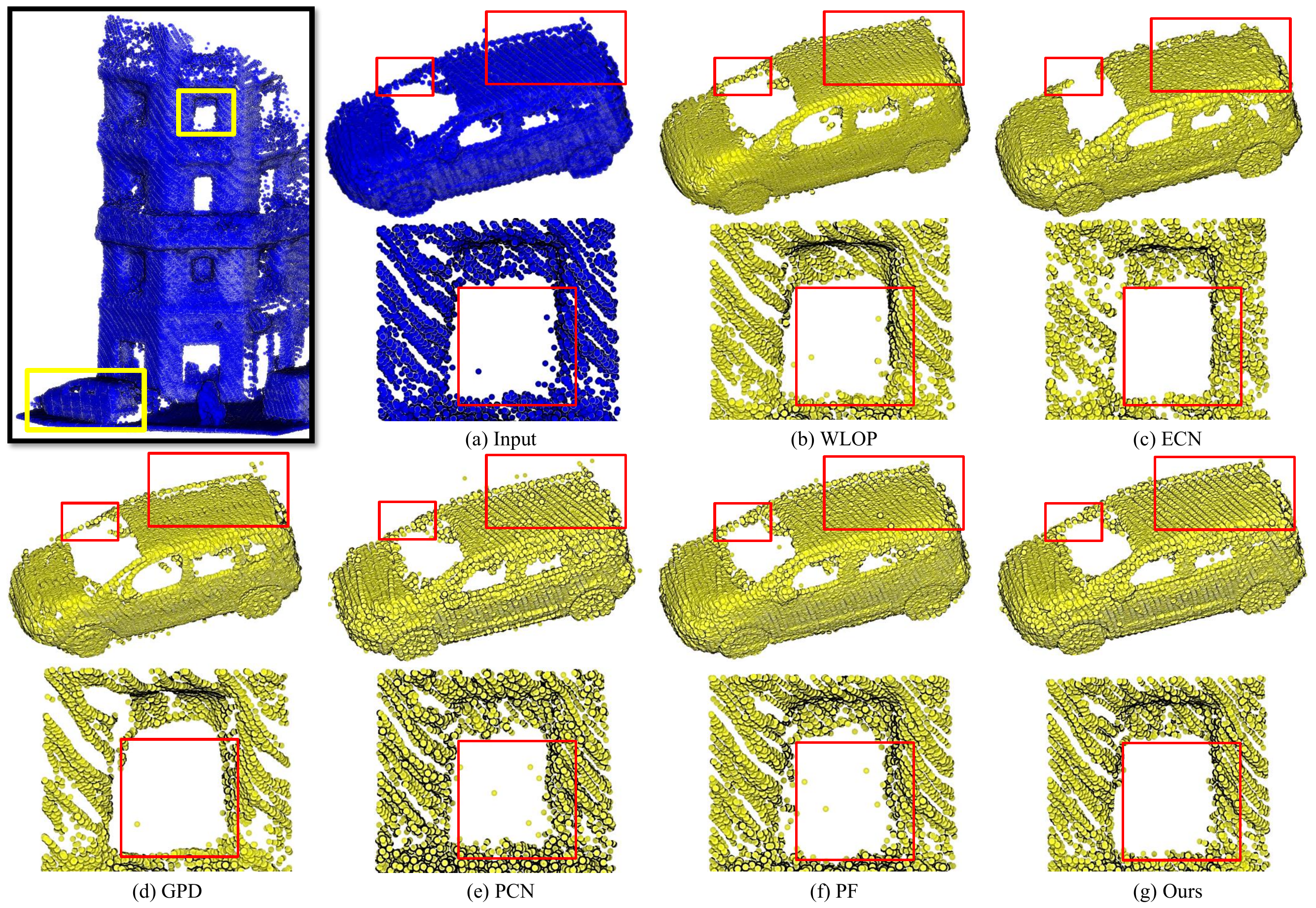}
  \caption{ \label{fig:raw_compare} Visual results of denoising methods on the real-world dataset Paris-rue-Madame.}
\end{figure*}

\subsection{Visual Comparison}

Besides the quantitative comparisons on synthetic dataset, we also show the visual comparisons on both synthetic noisy point clouds and raw-scan point clouds. Fig.~\ref{fig:Visual other methods} shows the denoising results from the proposed method and competitive baselines under 4 sets of $1.5\%$ Gaussian noisy point clouds. In order to test these methods on point clouds of different geometric complexity, these point clouds include both CAD models and non-CAD models.
From the results, the WLOP can not balance the denoising and the feature preservation. ECN and PCN have relatively weak denoising performance for point clouds with high-level noise. GPD tends to shape shrinkage. As a contrast, our MODNet significantly outperforms previous methods on both the simple parts point clouds and the complex curved surface point clouds. Our method not only achieves high performance denoising, but also does not lead to shape shrinkage and over-smooth geometric features.

As for raw-scan point clouds, we also verify the effectiveness of our method on the real-world dataset Paris-rue-Madame~\cite{serna2014paris} corrupted with raw noise. Since the ground truth models of these raw-scan point sets are not available, it may make the results difficult to objectively quantify. However, the performance of our method can still be partially demonstrated through visual comparison. And it is meaningful and essential to apply the proposed denoising method to raw scanned point clouds.
Therefore, we still demonstrate the visual comparisons (Fig.~\ref{fig:raw_compare}) with other methods, as suggested by previous techniques~\cite{zhang2020pointfilter,hermosilla2019total}. 

WLOP, ECN tend to oversmooth the results, As shown in Fig.~\ref{fig:raw_compare} (b, c). 
The shape of the point sets after GPD denoising are shrunk a bit. The PCN and PF seem not robust enough on raw-scan point clouds. As shown in Fig.~\ref{fig:raw_compare} (e, f), there is still a lot of noise after PCN or PF denoising. As a contrast, our MODNet can balance the denoising and the feature preservation well.

\begin{figure*}[!t]
  \centering
  \includegraphics[width=1\textwidth]{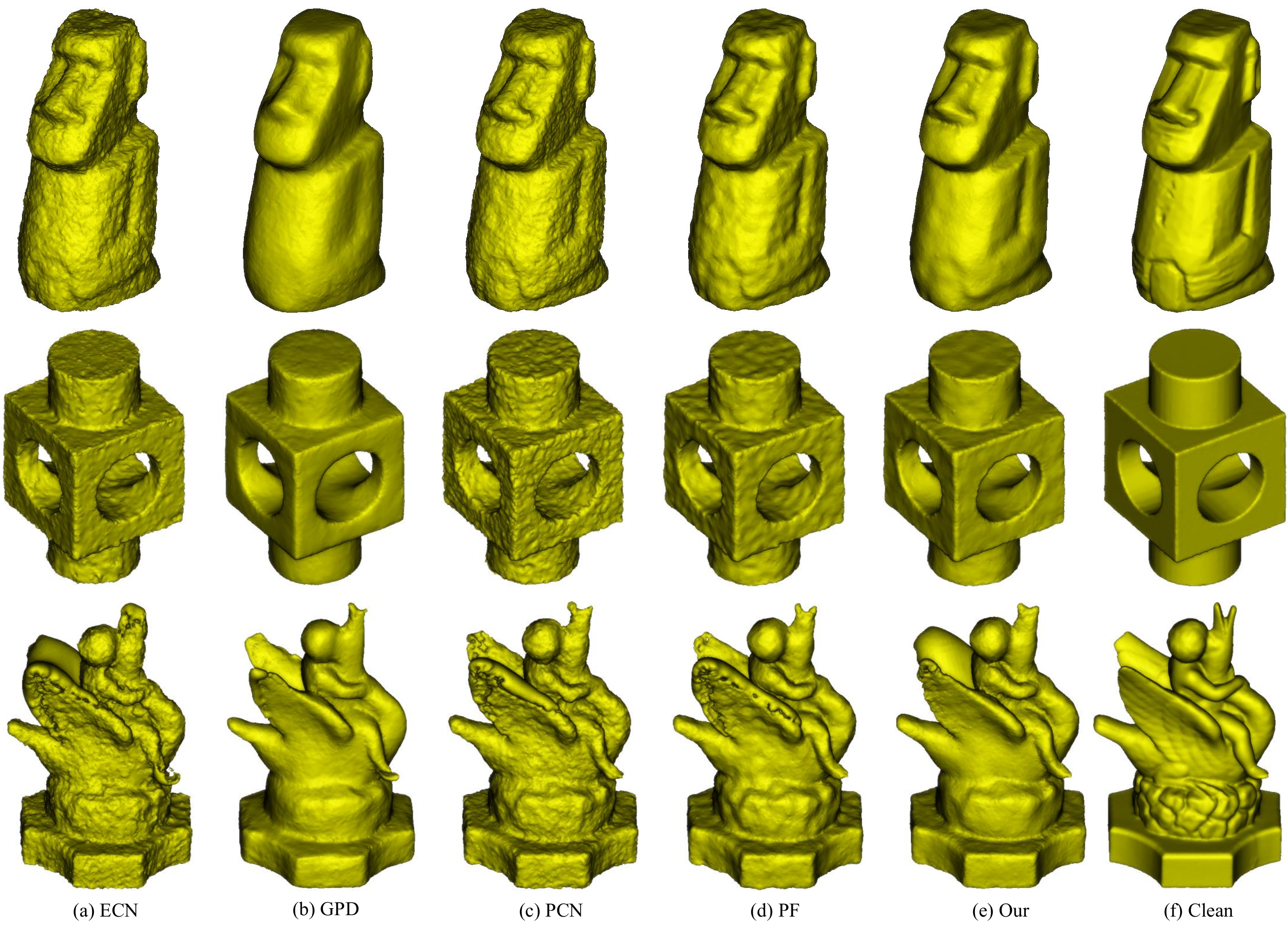}
  \caption{ \label{fig:reconstruction} Comparisons of Poisson reconstructed meshes of denoised results. From the Comparisons, the reconstructed meshes of the point cloud filtered by our method are cleaner and preserve more geometric details. }
\end{figure*}

In order to visualize the denoising results more intuitively, we reconstruct meshes from denoised point clouds by Poisson reconstruction. 
Fig.~\ref{fig:reconstruction} shows the meshes of denoising results from the proposed method and competitive baselines under 3 sets of $1.0\%$ Gaussian noisy point clouds.
Since the reconstructed surface of WLOP is poor, it is not shown. Only deep learning-based methods are compared here. From Fig.~\ref{fig:reconstruction} (b), although GPD outputs smoothing reconstructed meshes, it fails to preserve sharp features. 
The reconstructed mesh surface of ECN, PF, and PCN is relatively rough.
The reconstructed mesh surface of ECN is the roughest, even taking as input multi-scale patches. Therefore, we believe that simply aggregation multi-scale features is not necessarily effective in denoising task.
In contrast, the meshes of the point clouds filtered by our method are cleaner and preserve more geometric details.

\begin{figure}[!t]
  \centering
  \includegraphics[width=\linewidth]{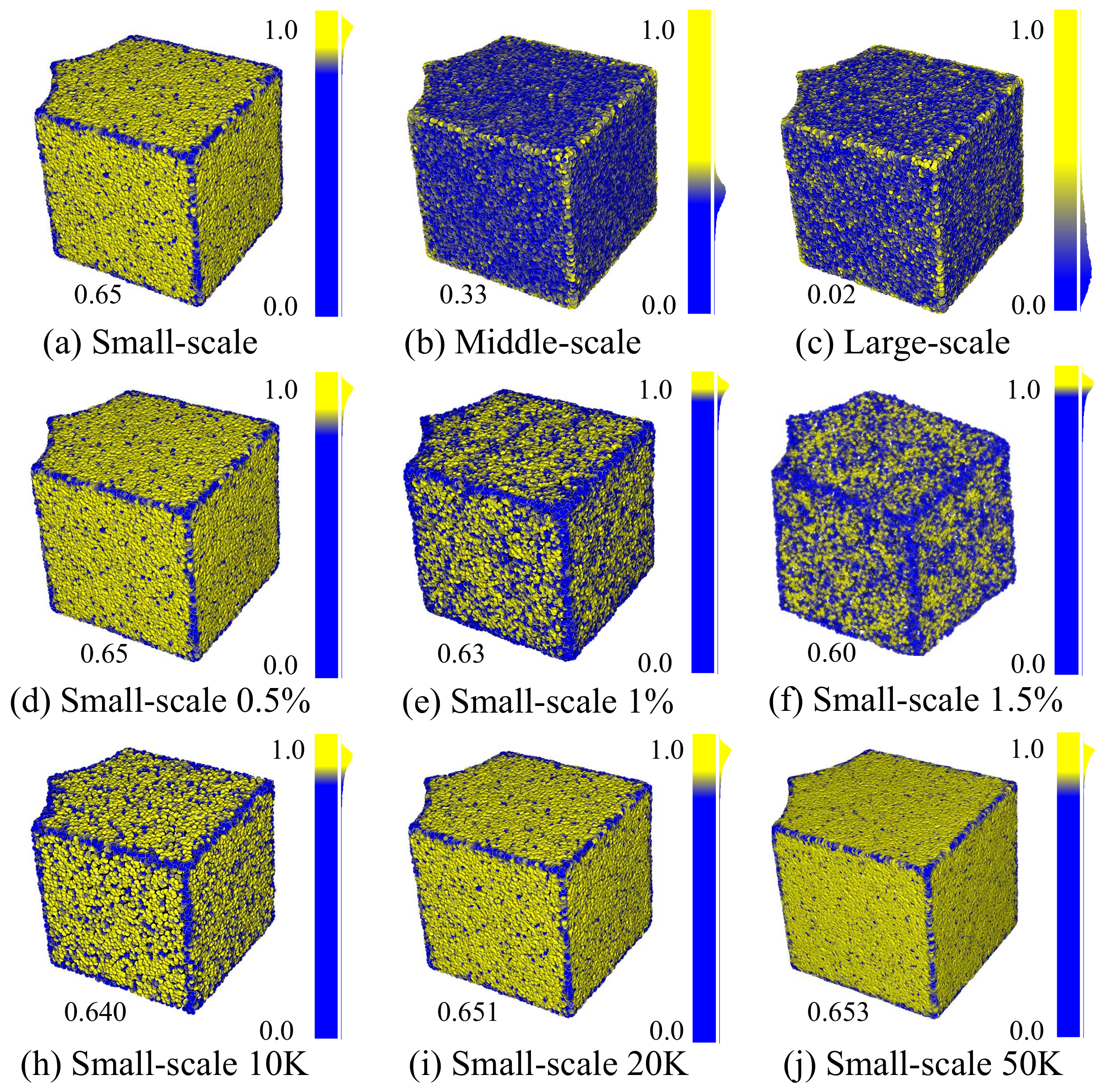}
  \caption{\label{fig:weights}Visual comparisons of the multi-scale weights. The models in the top row are three different scale weights, the models in the bottom two rows are the small-scale weights of models with different noise levels and densities. These numbers of the lower left corners of the models are the mean value of models' scale weights.}
\end{figure}

Furthermore, in order to demonstrate that MODNet can adaptively use the appropriate scale information according to the geometric information around noisy points, we sum the weights of the three offsets respectively. Then we visualize these weights. 
As shown in Fig.~\ref{fig:weights} (a,b,c), small-scale information has the greatest effect on denoising. For those edge points, MODNet tends to use more medium-scale and large-scale information. As for those points on plane, MODNet tends to use more small-scale information. 
This coincides with our priori knowledge. For points on simple geometric structures such as planes, small-scale information is sufficient. Middle-scale and large-scale information instead introduces redundant information. It sometimes even causes surface degradation.
For points on complex geometric structures such as edges, the addition of middle-scale and large-scale information leads to better denoising results.
As shown in Fig.~\ref{fig:weights} (d, e, f), the heigher noise level of point cloud require more middle-scale and large-scale information for denoising.
From the visualization results of Fig.~\ref{fig:weights} (g, h, i), higher density of point cloud need less middle-scale and large-scale information.
The above results also demonstrate the adaptive process of our MODNet for different point cloud noise levels and point cloud densities.

\begin{table}[]
\centering
\begin{tabular}{clcccc}
\hline
\multirow{2}{*}{Metrics} & \multicolumn{1}{c}{\multirow{2}{*}{Model}} & \multicolumn{4}{c}{Noise} \\ \cline{3-6} 
 & \multicolumn{1}{c}{} & DN & LN & UN & HLN \\ \hline
\multicolumn{1}{c|}{\multirow{6}{*}{CD$(10^{-4})$}} & WLOP & 8.81 & 9.38 & 12.5 & 15.2 \\
\multicolumn{1}{c|}{} & ECN & 5.31 & 6.12 & 5.84 & 12.9 \\
\multicolumn{1}{c|}{} & PCN & 3.45 & 3.79 & 7.12 & 12.7 \\
\multicolumn{1}{c|}{} & GPD & 3.42 & 3.89 & 4.25 & 12.1 \\
\multicolumn{1}{c|}{} & PF & 2.88 & 3.38 & 4.07 & 11.9 \\ \cline{2-6}
\multicolumn{1}{c|}{} & Ours & \textbf{2.71} & \textbf{3.19} & \textbf{3.19} & \textbf{7.99} \\ \hline\hline
\multicolumn{1}{c|}{\multirow{6}{*}{MSE$(10^{-2})$}} & WLOP & 4.43 & 4.35 & 3.41 &4.99 \\
\multicolumn{1}{c|}{} & ECN & 3.49 & 3.66 & 2.89 & 4.39 \\
\multicolumn{1}{c|}{} & PCN & 3.56 & 3.61 & 2.89 & 4.21 \\
\multicolumn{1}{c|}{} & GPD & 3.58 & 3.59 & 2.53 & 4.12 \\
\multicolumn{1}{c|}{} & PF & 3.40 & 3.42 & 3.47 & 4.01 \\ \cline{2-6}
\multicolumn{1}{c|}{} & Ours & \textbf{3.39} & \textbf{3.41} & \textbf{3.44} & \textbf{3.66} \\ \hline\hline
\multicolumn{1}{c|}{\multirow{6}{*}{P2M$(10^{-4})$}} & WLOP & 3.02 & 4.94 & 3.25 & 9.21 \\
\multicolumn{1}{c|}{} & ECN & 1.53 & 1.95 & 1.71 & 3.24 \\
\multicolumn{1}{c|}{} & PCN & 1.36 & 2.57 & 1.45 & 4.78 \\
\multicolumn{1}{c|}{} & GPD & 2.28 & 2.32 & 2.41 & 3.02 \\
\multicolumn{1}{c|}{} & PF & 0.81 & 1.31 & 0.83 & 4.77 \\ \cline{2-6}
\multicolumn{1}{c|}{} & Ours & \textbf{0.73} & \textbf{1.08} & \textbf{0.77} & \textbf{2.74} \\ \hline
\end{tabular}
\caption{\label{tab:Quantitative_rebust}Quantitative comparisons by testing various methods on synthetic noisy models with various noise models, including discrete noise (DN), Laplace noise (LN), uniform noise (UN) and 3$\%$ high-level Gaussian noise (HLN).}
\end{table}

\begin{table}[]
\centering
\begin{tabular}{c|ccc}
\hline
\multirow{2}{*}{$\beta$} & \multicolumn{3}{c}{Metrics}                                                                         \\ \cline{2-4} 
                   & \multicolumn{1}{c|}{CD$(10^{-4})$}            & \multicolumn{1}{c|}{MSE$(10^{-2})$}           & P2M$(10^{-4})$                      \\ \hline
0.0                  & \multicolumn{1}{c|}{2.82}          & \multicolumn{1}{c|}{2.49}          & 0.77                     \\
0.2                & \multicolumn{1}{c|}{\textbf{2.74}} & \multicolumn{1}{c|}{\textbf{2.48}} & \textbf{0.74}            \\
0.4                & \multicolumn{1}{c|}{2.86}          & \multicolumn{1}{c|}{2.49}          & 0.78                     \\
0.6                & \multicolumn{1}{c|}{2.81}          & \multicolumn{1}{c|}{2.49}          & 0.76                     \\
0.8                & \multicolumn{1}{c|}{2.81}          & \multicolumn{1}{c|}{2.49}          & 0.76                     \\ \hline\hline
\multirow{2}{*}{$K$} & \multicolumn{3}{c}{Metrics}                                                                         \\ \cline{2-4} 
                   & \multicolumn{1}{c|}{CD$(10^{-4})$}            & \multicolumn{1}{c|}{MSE$(10^{-2})$}           & P2M$(10^{-4})$                      \\ \hline
2                  & \multicolumn{1}{c|}{2.89}          & \multicolumn{1}{c|}{2.49}          & 0.80                     \\
3                  & \multicolumn{1}{c|}{\textbf{2.78}} & \multicolumn{1}{c|}{\textbf{2.48}} & \textbf{0.74}            \\
4                  & \multicolumn{1}{c|}{2.81}          & \multicolumn{1}{c|}{\textbf{2.48}} & \textbf{0.74}            \\
5                  & \multicolumn{1}{c|}{2.79}          & \multicolumn{1}{c|}{2.49}          & \textbf{0.74}            \\ \hline\hline
\multirow{2}{*}{$\left | \hat{p_{i,k}} \right |$} & \multicolumn{3}{c}{Metrics}                                                                         \\ \cline{2-4} 
                   & \multicolumn{1}{c|}{CD$(10^{-4})$}            & \multicolumn{1}{c|}{MSE$(10^{-2})$}           & P2M$(10^{-4})$                      \\ \hline
200                & \multicolumn{1}{c|}{2.82}          & \multicolumn{1}{c|}{2.49}          & 0.77                     \\
400                & \multicolumn{1}{c|}{\textbf{2.74}} & \multicolumn{1}{c|}{\textbf{2.48}} & \textbf{0.74}            \\
600                & \multicolumn{1}{c|}{2.87}          & \multicolumn{1}{c|}{2.49}          & 0.78                     \\
800                & \multicolumn{1}{c|}{2.90}          & \multicolumn{1}{c|}{2.50}          & 0.79 \\
1000               & \multicolumn{1}{c|}{2.93}          & \multicolumn{1}{c|}{2.51}          & 0.80                     \\ \hline\hline
\multirow{2}{*}{$r_{k}$} & \multicolumn{3}{c}{Metrics}                                                                         \\ \cline{2-4} 
                   & \multicolumn{1}{c|}{CD$(10^{-4})$}            & \multicolumn{1}{c|}{MSE$(10^{-2})$}           & P2M$(10^{-4})$                      \\ \hline
$2\%, 3\%, 4\%$                & \multicolumn{1}{c|}{2.94}          & \multicolumn{1}{c|}{2.49}          & 0.81                     \\
$3\%, 4\%, 5\%$                & \multicolumn{1}{c|}{\textbf{2.78}} & \multicolumn{1}{c|}{\textbf{2.48}} & \textbf{0.74}            \\
$4\%, 5\%, 6\%$                & \multicolumn{1}{c|}{2.79}          & \multicolumn{1}{c|}{2.50}          & \textbf{0.74}            \\
$5\%, 6\%, 7\%$                & \multicolumn{1}{c|}{\textbf{2.78}} & \multicolumn{1}{c|}{2.50}          & \textbf{0.74}            \\
$6\%, 7\%, 8\%$                & \multicolumn{1}{c|}{2.96}          & \multicolumn{1}{c|}{2.53}          & 0.79                     \\
$2\%, 5\%, 8\%$                & \multicolumn{1}{c|}{2.93} & \multicolumn{1}{c|}{2.50}          & 0.80            \\ 
$3\%, 5\%, 7\%$                & \multicolumn{1}{c|}{2.81} & \multicolumn{1}{c|}{2.49}          & 0.75            \\ \hline
\end{tabular}
 \caption{\label{tab:para}The comparisons with different hyper-parameter $\beta$, $K$, $\left | \hat{p_{i,k}} \right |$, $r_{k}$ on test set.}
\end{table}

\subsection{Hyperparameter selection}
In order to achieve better performance of our network, we do some comparative experiments on following hyperparameters, as shown in Table~\ref{tab:para}. The first hyperparameter $\beta$ is a trade-off parameter to balance $L_{dp}$ and $L_{p}^{final}$. When our MODNet is trained without $L_{dp}$ ($\beta=0$), it achieves worse denosing result than $\beta=0.2$. And the denoising performance will also decrease when our network pays too much attention to these three single scale denoising displacements($\beta>0.2$). According to the results of our experiments, we set $\beta=0.2$ in our training stage. $\left | \hat{p_{i,k}} \right |, r_{k}, K$ are hyperparameters of input patches. According to the results of our experiments, we set $\left | \hat{p_{i,k}} \right |=400, K=3 $, $r_{k}=\{3\%, 4\%, 5\%\}$ of the model’s bounding box diagonal length in our MODNet. In particular, we noticed that too large $\left | \hat{p_{i,k}} \right |$, $r_{k}$ and $K$ degrade the denoising performance instead. Therefore, selecting the optimal local neighborhood information is crucial for denoising, so it is a meaningful work to do in the future.

\subsection{Ablation Studies}



We further illustrate the effectiveness of different modules in MODNet, as shown in Table~\ref{tab:sub-module}. Pointfilter~\cite{zhang2020pointfilter} is employed as the baseline, and we analyze the effects of using different module settings on network performance. Based on PF, we add the remaining modules back one at a time to obtain a new model, which is denoted as M1 to M4 respectively. M4 is our full pipeline with all the modules. First, Using the PFE module alone gains lower CD, MSE and P2M, compared with the Pointfilter. 
Compared with the single-scale patch, multi-scale patches information leads to a notable performance improvement of denoising. Based on M1, we add the MOD module to obtain model M2. The MOD of M2 uses as input the output of PFE, and the multi-scale weights are set to an all-one matrix. 
And the MOD of M2 doesn't predicts  $dp_{i,k}^{pre}$. 
As for M3, we add the MSPM module in front of the MOD, and the multi-scale weights were acquired by MSPM. 
M4* doesn't predicts $dp_{i,k}^{pre}$ like M2, but directly supervises $offset_{i,k}$ with $L_{dp}$. 
M4 predicts $dp_{i,k}^{pre}$ and employs the loss $L_{dp}$ to supervise $dp_{i,k}^{pre}$. 
From the results, directly supervising $offset_{i,k}$ seems to limit the learning ability of the network and instead leads to denoising performance degradation.
Therefore, PFE, MSPM, MOD and $L_{dp}$ all contribute positively to the performance.

\begin{table}[]
\centering
\begin{tabular}{lcccc|ccc}
\hline
\multirow{2}{*}{Model} & \multirow{2}{*}{PFE} & \multirow{2}{*}{MOD} & \multirow{2}{*}{MSPM} & \multirow{2}{*}{$L_{dp}$} & \multicolumn{3}{c}{Metrics} \\ \cline{6-8} 
   &   &   &   &   & \multicolumn{1}{c|}{CD}   & \multicolumn{1}{c|}{MSE}  & P2M  \\ \hline
PF &   &   &   &   & \multicolumn{1}{c|}{3.16} & \multicolumn{1}{c|}{2.52} & 0.89 \\
M1 & \checkmark &   &   &   & \multicolumn{1}{c|}{2.92} & \multicolumn{1}{c|}{2.50} & 0.80 \\
M2 & \checkmark & \checkmark &   &   & \multicolumn{1}{c|}{2.85} & \multicolumn{1}{c|}{2.49} & 0.77 \\
M3 & \checkmark & \checkmark & \checkmark &   & \multicolumn{1}{c|}{2.79} & \multicolumn{1}{c|}{2.49} & 0.77 \\
M4* & \checkmark & \checkmark & \checkmark & \checkmark & \multicolumn{1}{c|}{2.80} & \multicolumn{1}{c|}{2.49} & 0.75\\ \hline
M4 & \checkmark & \checkmark & \checkmark & \checkmark & \multicolumn{1}{c|}{\textbf{2.78}} & \multicolumn{1}{c|}{\textbf{2.48}} & \textbf{0.74} \\ \hline
\end{tabular}
\caption{\label{tab:sub-module}Effectiveness of different sub-module of our model. CD is multiplied by $10^{4}$, MSE is multiplied by $10^{2}$ and P2M is multiplied by $10^{4}$. }
\end{table}

\section{Conclusion}
\label{conclusion}


In this paper, we propose a multi-offset denoising network customized for multi-scale patches input, which employs multi-scale geometric perception information to guide the network to utilize multi-scale information. It is a semi-soft neighborhood selecting mechanism. 
More specifically, the MODNet is employed to extract the features of multi-scale patches separately. And it regresses three different scale denoising offsets.  
And three different scale denoising offsets are guided by the multi-scale weights to predict the final denoising displacement by weighting them adaptively. 
A variety of experiments demonstrate that our method can yield state-of-the-art denoising performance.

\textbf{Limitation.} 
Our method is a new paradigm for using multi-scale patches information on denoising networks. 
It is a semi-soft neighborhood selecting mechanism. However, the number of scales and patch scale are still hyper-parameters that need to be determined. And the running time of our MODNet is not very fast.

\section*{Acknowledgments}
\label{acknowledgments}

This work was supported by the National Key Research and Development Program of China (No. 2020YFB2010702, No. 2019YFB1707504, 2019YFB1707501) and Natural Science Foundation of Jiangsu Province (No.BK20190016).


\bibliographystyle{eg-alpha-doi} 
\bibliography{modnet}       

\newcommand{\etalchar}[1]{$^{#1}$}
\begin{thebibliography}{\uppercase{LCOLTE07}}

\bibitem[ABCO{\etalchar{*}}01]{alexa2001point}
\textsc{Alexa M., Behr J., Cohen-Or D., Fleishman S., Levin D., Silva C.~T.}:
\newblock Point set surfaces.
\newblock In \emph{Proceedings Visualization, 2001. VIS'01.} (2001), IEEE,
  pp.~21--29.

\bibitem[ABCO{\etalchar{*}}03]{alexa2003computing}
\textsc{Alexa M., Behr J., Cohen-Or D., Fleishman S., Levin D., Silva C.~T.}:
\newblock Computing and rendering point set surfaces.
\newblock \emph{TVCG 9}, 1 (2003), 3--15.

\bibitem[ASGCO10]{avron2010}
\textsc{Avron H., Sharf A., Greif C., Cohen-Or D.}:
\newblock L1-sparse reconstruction of sharp point set surfaces.
\newblock \emph{TOG 29}, 5 (2010), 1--12.

\bibitem[CP05]{cazals2005estimating}
\textsc{Cazals F., Pouget M.}:
\newblock Estimating differential quantities using polynomial fitting of
  osculating jets.
\newblock \emph{CAGD 22}, 2 (2005), 121--146.

\bibitem[GG07]{2007Algebraic}
\textsc{Guennebaud G., Gross M.}:
\newblock Algebraic point set surfaces.
\newblock \emph{Acm Transactions on Graphics 26}, 3 (2007),
  23--1---23--23--923--23--9.

\bibitem[GKOM18]{guerrero2018pcpnet}
\textsc{Guerrero P., Kleiman Y., Ovsjanikov M., Mitra N.~J.}:
\newblock Pcpnet learning local shape properties from raw point clouds.
\newblock In \emph{CGF} (2018), vol.~37, Wiley Online Library, pp.~75--85.

\bibitem[HGCG20]{hu2020feature}
\textsc{Hu W., Gao X., Cheung G., Guo Z.}:
\newblock Feature graph learning for 3d point cloud denoising.
\newblock \emph{IEEE Transactions on Signal Processing 68} (2020), 2841--2856.

\bibitem[HHWG21]{hu2021dynamic}
\textsc{Hu W., Hu Q., Wang Z., Gao X.}:
\newblock Dynamic point cloud denoising via manifold-to-manifold distance.
\newblock \emph{IEEE TIP 30} (2021), 6168--6183.

\bibitem[HLZ{\etalchar{*}}09]{huang2009consolidation}
\textsc{Huang H., Li D., Zhang H., Ascher U., Cohen-Or D.}:
\newblock Consolidation of unorganized point clouds for surface reconstruction.
\newblock \emph{TOG 28}, 5 (2009), 1--7.

\bibitem[HPL{\etalchar{*}}21]{hu2021graph}
\textsc{Hu W., Pang J., Liu X., Tian D., Lin C.-W., Vetro A.}:
\newblock Graph signal processing for geometric data and beyond: Theory and
  applications.
\newblock \emph{IEEE Transactions on Multimedia} (2021).

\bibitem[HRR19]{hermosilla2019total}
\textsc{Hermosilla P., Ritschel T., Ropinski T.}:
\newblock Total denoising: Unsupervised learning of 3d point cloud cleaning.
\newblock In \emph{ICCV} (2019), pp.~52--60.

\bibitem[HSH{\etalchar{*}}20]{he2020pvn3d}
\textsc{He Y., Sun W., Huang H., Liu J., Fan H., Sun J.}:
\newblock Pvn3d: A deep point-wise 3d keypoints voting network for 6dof pose
  estimation.
\newblock In \emph{CVPR} (2020), pp.~11632--11641.

\bibitem[HWG{\etalchar{*}}13]{huang2013edge}
\textsc{Huang H., Wu S., Gong M., Cohen-Or D., Ascher U., Zhang H.}:
\newblock Edge-aware point set resampling.
\newblock \emph{TOG 32}, 1 (2013), 1--12.

\bibitem[LCOLTE07]{lipman2007parameterization}
\textsc{Lipman Y., Cohen-Or D., Levin D., Tal-Ezer H.}:
\newblock Parameterization-free projection for geometry reconstruction.
\newblock \emph{TOG 26}, 3 (2007), 22--es.

\bibitem[LWC{\etalchar{*}}17]{lu2017gpf}
\textsc{Lu X., Wu S., Chen H., Yeung S.-K., Chen W., Zwicker M.}:
\newblock Gpf: Gmm-inspired feature-preserving point set filtering.
\newblock \emph{TVCG 24}, 8 (2017), 2315--2326.

\bibitem[LXW{\etalchar{*}}22]{lu2022transformers}
\textsc{Lu D., Xie Q., Wei M., Xu L., Li J.}:
\newblock Transformers in 3d point clouds: A survey.
\newblock \emph{arXiv preprint arXiv:2205.07417} (2022).

\bibitem[MC17]{mattei2017point}
\textsc{Mattei E., Castrodad A.}:
\newblock Point cloud denoising via moving rpca.
\newblock In \emph{CGF} (2017), vol.~36, Wiley Online Library, pp.~123--137.

\bibitem[NIL12]{nguyen2012modeling}
\textsc{Nguyen C.~V., Izadi S., Lovell D.}:
\newblock Modeling kinect sensor noise for improved 3d reconstruction and
  tracking.
\newblock In \emph{2012 second international conference on 3D imaging,
  modeling, processing, visualization \& transmission} (2012), IEEE,
  pp.~524--530.

\bibitem[{\"O}GG09]{oztireli2009feature}
\textsc{{\"O}ztireli A.~C., Guennebaud G., Gross M.}:
\newblock Feature preserving point set surfaces based on non-linear kernel
  regression.
\newblock In \emph{CGF} (2009), vol.~28, Wiley Online Library, pp.~493--501.

\bibitem[PFVM20]{pistilli2020learning}
\textsc{Pistilli F., Fracastoro G., Valsesia D., Magli E.}:
\newblock Learning graph-convolutional representations for point cloud
  denoising.
\newblock In \emph{ECCV} (2020), Springer, pp.~103--118.

\bibitem[QFC{\etalchar{*}}20]{qi2020p2b}
\textsc{Qi H., Feng C., Cao Z., Zhao F., Xiao Y.}:
\newblock P2b: Point-to-box network for 3d object tracking in point clouds.
\newblock In \emph{CVPR} (2020), pp.~6329--6338.

\bibitem[QSMG17]{2017PointNet}
\textsc{Qi C.~R., Su H., Mo K., Guibas L.~J.}:
\newblock Pointnet: Deep learning on point sets for 3d classification and
  segmentation.
\newblock In \emph{CVPR} (2017).

\bibitem[QYSG17]{qi2017pointnet++}
\textsc{Qi C.~R., Yi L., Su H., Guibas L.~J.}:
\newblock Pointnet++: Deep hierarchical feature learning on point sets in a
  metric space.
\newblock \emph{arXiv preprint arXiv:1706.02413} (2017).

\bibitem[RLBG{\etalchar{*}}20]{rakotosaona2020pointcleannet}
\textsc{Rakotosaona M.-J., La~Barbera V., Guerrero P., Mitra N.~J., Ovsjanikov
  M.}:
\newblock Pointcleannet: Learning to denoise and remove outliers from dense
  point clouds.
\newblock In \emph{CGF} (2020), vol.~39, Wiley Online Library, pp.~185--203.

\bibitem[R{\"O}PG18]{roveri2018pointpronets}
\textsc{Roveri R., {\"O}ztireli A.~C., Pandele I., Gross M.}:
\newblock Pointpronets: Consolidation of point clouds with convolutional neural
  networks.
\newblock In \emph{CGF} (2018), vol.~37, Wiley Online Library, pp.~87--99.

\bibitem[RRN{\etalchar{*}}20]{ravi2020accelerating}
\textsc{Ravi N., Reizenstein J., Novotny D., Gordon T., Lo W.-Y., Johnson J.,
  Gkioxari G.}:
\newblock Accelerating 3d deep learning with pytorch3d.
\newblock \emph{arXiv preprint arXiv:2007.08501} (2020).

\bibitem[SMGD14]{serna2014paris}
\textsc{Serna A., Marcotegui B., Goulette F., Deschaud J.-E.}:
\newblock Paris-rue-madame database: a 3d mobile laser scanner dataset for
  benchmarking urban detection, segmentation and classification methods.
\newblock In \emph{ICPRAM} (2014).

\bibitem[SNF{\etalchar{*}}13]{shuman2013emerging}
\textsc{Shuman D.~I., Narang S.~K., Frossard P., Ortega A., Vandergheynst P.}:
\newblock The emerging field of signal processing on graphs: Extending
  high-dimensional data analysis to networks and other irregular domains.
\newblock \emph{IEEE signal processing magazine 30}, 3 (2013), 83--98.

\bibitem[SPV15]{schoenenberger2015graph}
\textsc{Schoenenberger Y., Paratte J., Vandergheynst P.}:
\newblock Graph-based denoising for time-varying point clouds.
\newblock In \emph{2015 3DTV-Conference: The True Vision-Capture, Transmission
  and Display of 3D Video (3DTV-CON)} (2015), IEEE, pp.~1--4.

\bibitem[SSW15]{sun2015denoising}
\textsc{Sun Y., Schaefer S., Wang W.}:
\newblock Denoising point sets via l0 minimization.
\newblock \emph{CAGD 35} (2015), 2--15.

\bibitem[TQD{\etalchar{*}}19]{thomas2019kpconv}
\textsc{Thomas H., Qi C.~R., Deschaud J.-E., Marcotegui B., Goulette F., Guibas
  L.~J.}:
\newblock Kpconv: Flexible and deformable convolution for point clouds.
\newblock In \emph{ICCV} (2019), pp.~6411--6420.

\bibitem[XLH{\etalchar{*}}20]{xie2020rrcnet}
\textsc{Xie Q., Lu D., Huang A., Yang J., Li D., Zhang Y., Wang J.}:
\newblock Rrcnet: Rivet region classification network for rivet flush
  measurement based on 3-d point cloud.
\newblock \emph{IEEE Transactions on Instrumentation and Measurement 70}
  (2020), 1--12.

\bibitem[YLF{\etalchar{*}}18]{yu2018ec}
\textsc{Yu L., Li X., Fu C.-W., Cohen-Or D., Heng P.-A.}:
\newblock Ec-net: an edge-aware point set consolidation network.
\newblock In \emph{ECCV} (2018), pp.~386--402.

\bibitem[ZCN{\etalchar{*}}19]{zeng20193d}
\textsc{Zeng J., Cheung G., Ng M., Pang J., Yang C.}:
\newblock 3d point cloud denoising using graph laplacian regularization of a
  low dimensional manifold model.
\newblock \emph{IEEE TIP 29} (2019), 3474--3489.

\bibitem[ZLQH20]{zhang2020pointfilter}
\textsc{Zhang D., Lu X., Qin H., He Y.}:
\newblock Pointfilter: Point cloud filtering via encoder-decoder modeling.
\newblock \emph{TVCG 27}, 3 (2020), 2015--2027.

\end{thebibliography}


\newpage

\end{document}